\documentclass[letterpaper, 10 pt, conference]{ieeeconf}  

\IEEEoverridecommandlockouts                              

\usepackage{graphics} 
\usepackage{epsfig} 
\usepackage{hyperref}
\usepackage{cleveref}
\crefname{figure}{Fig.}{Figs.}
\Crefname{figure}{Fig.}{Figs.}
\crefname{table}{Table}{Tables}
\Crefname{table}{Table}{Tables}
\usepackage{mathrsfs}
\usepackage{tabularx}
\usepackage{booktabs}

\title{\LARGE \bf
SASI: Leveraging Sub-Action Semantics for Robust Early Action Recognition in Human-Robot Interaction
}
\author{
Yongpeng Cao$^{1, *}$\thanks{$^{*}$Corresponding author}%
\thanks{$^{1}$Institute of Industrial Science, The University of Tokyo, Japan}
, Masahiro Hirano$^{1}$, Hyuno Kim$^{1}$, and Yuji Yamakawa$^{2}$%
\thanks{$^{2}$Interfaculty Initiative in Information Studies, Graduate School of Interdisciplinary Information Studies, The University of Tokyo, Japan}%
\\ \{cao, mhirano, h-kim, y-ymkw\}@iis.u-tokyo.ac.jp
}
\begin{document}

\maketitle
\thispagestyle{empty}
\pagestyle{empty}

\begin{abstract}
Understanding human actions is critical for advancing behavior analysis in human–robot interaction. Particularly in tasks that demand quick and proactive feedback, robots must recognize human actions as early as possible from incomplete observations. \textit{Sub-actions} offer the semantic and hierarchical cues needed for this, since human actions are inherently structured and can be decomposed into smaller, meaningful units. However, conventional approaches focus primarily on holistic actions and often overlook the rich semantic structure embedded in sub-actions, making them poorly suited for early recognition. To address this gap, we introduce SASI (Sub-Action Semantics Integrated cross-modal fusion), a novel framework that integrates existing graph convolution networks to fuse spatiotemporal features with sub-action semantics. SASI exploits a segmentation model with a traditional skeleton-based graph convolution network, capturing both fine-grained sub-action semantics and overall spatial context, while operating in real-time at 29 Hz. Experiments on BABEL, a skeleton-based dataset with frame-level annotations, demonstrate that our method improves recognition accuracy over conventional approaches, with additional gains expected as the quality of sub-action segmentation improves. Notably, SASI also achieves superior performance in understanding partial action sequences, revealing its capability for early recognition, which is essential for proactive and seamless Human-Robot Interaction (HRI). Code is available at \href{https://github.com/SavickTso/SASI}{https://github.com/SavickTso/SASI} . 
\end{abstract}

\section{Introduction}
\begin{figure}[htp!]
  \centering
  \setlength{\abovecaptionskip}{2pt}
  \includegraphics[width=\linewidth]{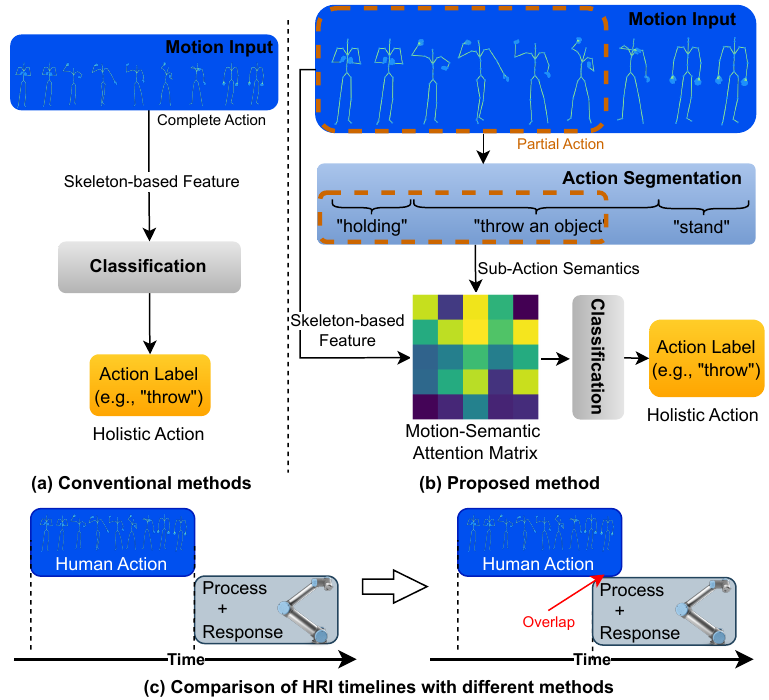}
  \caption{A comparison between the conventional holistic approach and proposed method for action recognition. (a) Conventional holistic approaches directly map spatiotemporal skeleton-based features to a final classification. (b) Our method explicitly models the compositional nature of actions. We segment the motion into sub-actions, which are then encoded as text embeddings. A cross-attention mechanism fuses motion and text modalities, aligning fine-grained sub-actions (text) with kinematic patterns (motion) to enhance holistic action recognition. (c) Compared to conventional methods, our approach is capable of introducing timeline overlap into the human-robot interaction workflow, enabling proactive feedback.}
  \label{fig:concept}
\end{figure}
Human action understanding, particularly action recognition, plays a pivotal role in applications such as human-robot collaboration and intelligent industrial systems. While graph convolution networks (GCNs) on human skeletons have advanced recognition performance~\cite{GCN, stgcn}, two key challenges remain for diverse tasks and real-time collaboration: \textit{action complexity}, and \textit{early action recognition}.

The action complexity arises from the inherent hierarchical structure of human actions. Conventional methods often extract only coarse spatiotemporal features, which results in representations that fail to capture the subtle, hierarchical nuances between different actions. This simplification can limit the overall interpretability and the recognition performance. Early action recognition is c
ommon in real-world scenarios where systems, such as assistive robots, are required to work in real-time. These systems often receive only partial motion sequences, yet most existing action recognition methods are designed under the assumption that complete action data is available. Beyond the latency introduced by waiting for entire action sequences, this mismatch can cause incorrect recognition, leading to failed interactions in HRI tasks.

To tackle these problems, we propose a novel solution inspired by human action nature. Human actions naturally exhibit a hierarchical structure in which complex actions are decomposable into a series of motion primitives~\cite{lan2015action}. For instance, as shown in~\cref{fig:concept}, the sub-action ``throw an object'' serves as a key unit in the action ``throw''. This nature enables humans to infer the general intended behavior. Modeling such a hierarchical structure improves both accuracy and robustness when dealing with partially observed sequences. Moreover, capturing the temporal arrangement of sub-actions allows for early prediction of actions. For instance, we can predict the next move or the holistic action by observing the sub-action ``open the bottle cap'', which is possibly executed before ``drinking'' or bottle-related actions. This feature is especially critical for real-time applications such as Human-Robot Interaction. Previous studies also demonstrate that motion decomposition and the explicit modeling of motion primitives such as directional motion of limb can significantly boost performance in action recognition tasks~\cite{maerecognition, motionprimitives}.

To seek the promising possibility of utilizing the decomposable nature of human actions for action recognition, we introduce sub-action semantics integrated cross-modal fusion (SASI). SASI is a bottom-up pluggable framework designed to augment standard GCN backbones. It unifies holistic action classification with sub-action segmentation-based action semantic analysis, where we develop a \textit{sub-action semantics branch} on top of the action segmentation model to capture the correlation between sub-actions and holistic actions. This branch works along with the GCN-based \textit{action kinematic branch} to fuse semantic features and the skeleton-based features together for the final recognition. The schematic diagram of the proposed method is shown in~\cref{fig:concept} (b). Compared to prior works, SASI models the hierarchical relationship between holistic action (e.g., ``throw'') and sub-actions (e.g., ``stand'', ``throw an object'', ``holding''), allowing for robust recognition even from incomplete sequences—a capability essential for seamless real-time HRI.

Our main contributions in this work are:
\begin{itemize}
\item A cross-modal fusion architecture that combines skeleton-based features and sub-action semantics. 
\item A bottom-up sub-action semantics branch that leverages a pre-trained segmentation model to model hierarchical dependencies between holistic actions and sub-actions, enhancing robustness to incomplete motion sequences common in real-time HRI.
\item Experiments on the BABEL benchmark demonstrating the effectiveness of the proposed method and its potential to reduce response latency in collaborative tasks through early action recognition.
\end{itemize}

\section{Related Work}
\subsection{Skeleton-based Human Action Recognition}

Skeleton-based action recognition has traditionally relied on motion capture (MoCap) datasets such as Human3.6M~\cite{human36m} to record precise joint positions. Early methods employ classical classifiers and recurrent neural networks (RNN) to model the temporal dynamics within joint sequences, as demonstrated by Du et al. \cite{du2015hierarchical}. Subsequent studies~\cite{liu2017skeleton, song2018spatio} also adopt long short-term memory (LSTM) networks to effectively capture both spatial and temporal features inherent in motion sequences. Further progress is achieved by applying convolutional neural networks (CNNs) to skeletal data, where the sequences are transformed into image-like representations, as illustrated in the work of Du et al. \cite{du2015cnn}. This insight lays the groundwork for graph convolutional networks (GCNs), which naturally model the human skeleton as a graph with joints as nodes and the physical connections between them as edges. In this context, Yan et al. \cite{yan2018spatial} propose Spatial Temporal Graph Convolutional Networks (ST-GCN) to capture spatial and temporal dependencies simultaneously, which significantly improves the human action recognition performance. Later approaches improved upon this paradigm by learning adaptive graph topologies that relax the strict constraints of conventional graph convolutions~\cite{CTR-GCN, cheng2020skeleton, degcn, blockgcn, protoGCN, chi2022infogcn, 2sagcn2019cvpr, liu2023temporal, infogcn++}. Although powerful, these architectures typically fail to model the compositional structure required for fine-grained understanding.

\subsection{Multi-modal Human Action Understanding}
Building on advances in action estimation, datasets such as NTU-RGB+D~\cite{shahroudy2016ntu} extend the analysis to RGB video-based action understanding by providing the image data as an additional modality. Recent works also explore the integration of contextual cues, such as human-environment interactions~\cite{mao2022contact}, gaze information in human-robot interaction~\cite{gazetrack}, and depth cues from 3D data~\cite{depth3dcnn, PoseConv3D}. Unlike conventional approaches, our method focuses on integrating sub-action semantics with skeleton-based features to enhance interpretability and robustness. Zhang et al. \cite{zhang2024pevl} utilize pre-trained vision-language models (VLMs) for contrastive learning, which enhances the alignment between visual features and textual semantics. Despite these advancements, the hierarchical semantics in complex human actions remains under-explored.
\subsection{Early Action Prediction}
Early action prediction is critical for real-time systems, such as assistive robotics, as it involves inferring ongoing actions from partial observations. Methods such as~\cite{kendo} utilize prior knowledge to concentrate on key segments of an action, thus enabling early and effective prediction. \cite{gazetrack} leverages gaze information for human intention prediction. ~\cite{stergiou2023wisdom} use attention to weight discriminative early motion frames, while~\cite{liu2023rich} models action-semantic consistent knowledge. However, these techniques often fall short in considering sub-action semantic grounding, which limits their interpretability and stability. 
\subsection{Action Recognition For Human-Robot Interaction}

Action recognition enables robots to anticipate and adapt to human behaviors in HRI. Early studies explored multi-modal approaches, such as combining speech and vision, to improve recognition robustness in assistive interactions \cite{rodomagoulakis2016multimodal}. For coordination, methods such as trajectory mapping between humans and robots \cite{maeda2017probabilistic} and online human motion prediction \cite{butepage2017anticipating} allow robots to operate with awareness of human actions. Trajectory-based human intent prediction further supports safe, collision-free collaboration \cite{lyu2022efficient}. Moreover, action-conditioned interaction has been explored by transferring human–human interaction knowledge to human-robot interactions for an efficient learning process and more responsive robot behaviors \cite{kedia2024interact}. On the other hand, teleoperation-based learning frameworks such as Mobile ALOHA \cite{fu2024mobile} enable motion cloning for practical HRI tasks. Across these applications, action recognition serves not only as a perception module but also as a foundation for intent inference, fluent collaboration, and safety in real-world HRI. Nevertheless, prior approaches rarely model the semantic structure of actions at the sub-action level. 
\medskip 

In summary, a critical gap exists at the intersection of skeleton-based human action recognition and HRI: the explicit modeling of hierarchical semantic structure for robust early recognition in real-time robotics. Our work addresses this gap by integrating sub-action semantics into a GCN-based recognition pipeline.

\begin{figure*}[thp!]
  \vspace{2mm}
  \centering
  \setlength{\abovecaptionskip}{2pt} 
  \includegraphics[width=0.89\linewidth]{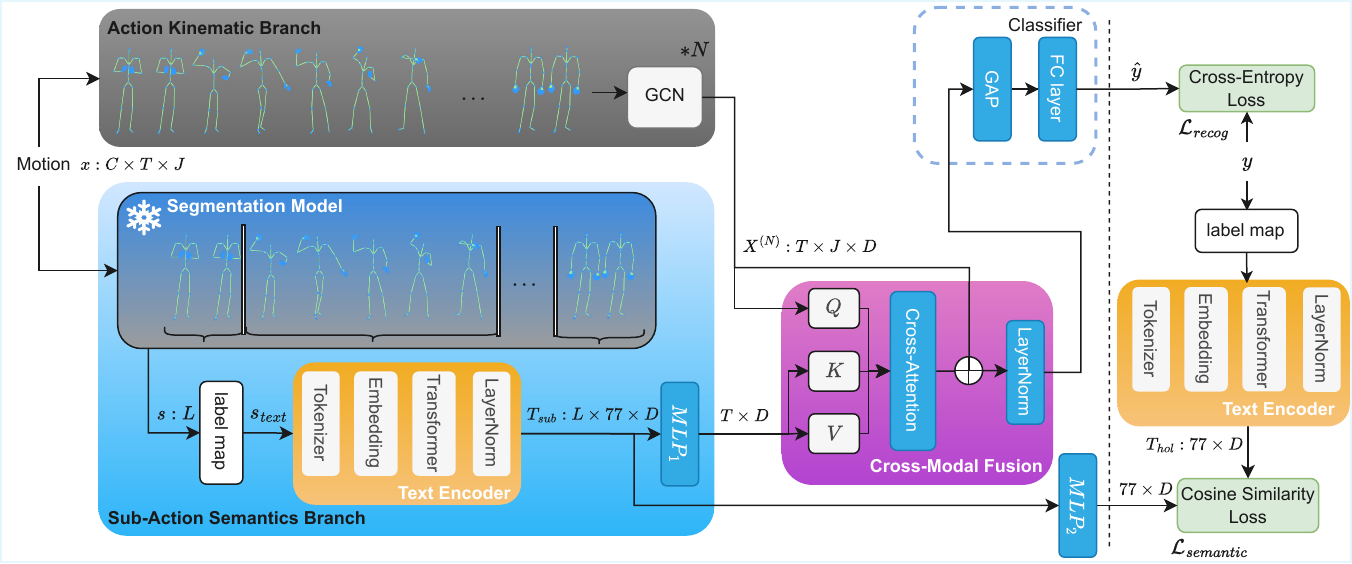}
  \caption{Framework overview of the proposed method. The MoCap motion data $x$ is concurrently processed by two parallel branches: an action kinematic branch, which extracts spatiotemporal $X^{(N)}$ using a GCN-based architecture, and a sub-action semantics branch, which generates sub-action embeddings $T_{sub}$ using a pre-trained segmentation model and a text encoder. The outputs are aligned by a cross-model fusion for classification. The network is jointly refined by using both recognition loss and semantic loss. The tensor dimensions are denoted as follows: $C$ is the channel dimension, $T$ is the temporal length, $J$ is the number of joints, $D$ is the feature dimension, and $L$ is the number of sub-actions. The context length of the text encoder is fixed at 77.}
  \label{fig:arch}
\end{figure*}
\section{Methodology}

Most conventional human action recognition methods treat actions as monolithic, continuous sequences. Therefore they usually fail to capture the complexity of real-world behaviors. However, inspired by the FineGym~\cite{shao2020finegym} and BABEL~\cite{BABEL:CVPR:2021}, we notice that human actions are decomposable into sub-actions with implicit semantic connections to their parent actions. Therefore, to leverage this semantic relationship and improve the general recognition accuracy, we adopt the dual-branch architecture to fuse the sub-action semantic features with spatiotemporal features for skeleton-based action recognition. 

First, the action kinematic branch adopts the skeleton GCN architecture for skeleton-based feature extraction, where we pass the raw motion input to GCN blocks. Then, a sub-action semantics branch is constructed to segment action by a pre-trained action segmentation model. Segmented sub-action classes are passed to a text encoder to extract sub-action semantics. In the end, we apply the cross-attention-based cross-modal fusion to skeleton-based features and sub-action semantics for the final classification. In addition, a semantic loss between the holistic action and its sub-actions is computed to update the sub-action semantics branch for hierarchical action understanding. This is jointly optimized with the action recognition loss to refine the overall network. The framework is shown in~\cref{fig:arch}

\subsection{GCN Backbone}
GCNs are usually used to model anatomical joint relationships through learnable adjacency matrices. We keep the GCN-based network as the backbone to extract kinematic features from raw human motions. Similar to conventional methods~\cite{2sagcn2019cvpr, CTR-GCN}, we use $N$ GCN-based blocks to stabilize and refine the features. 
Given the raw motion data $x$, we pass it through a series of GCN layers to extract spatiotemporal skeleton-based features. At the $l$-th layer, the skeleton-based features $X^{(l)}$ are computed as:
\begin{equation}
X^{(l)} = \sigma(A X^{(l-1)}W^{(l)}),
\end{equation}
where $X^{(0)}$ is $x$, and $W^{(l)}$ is the weights of GCN layer $l$. $A$ is the adjacency matrix representing human joint connections that are learnable to adapt to different actions. After processing through $N$ GCN layers, we get the final skeleton-based features as $X^{(N)}$.

\subsection{Action Segmentation and Sub-Action Semantics Branch}
To extract semantic information from sub-actions, we design the sub-action semantics branch utilizing an action segmentation model to recognize sub-actions from the motion sequence. The segmentation model is pre-trained on the same dataset as the primary network to keep the continuity. During the training of the primary network, pre-trained segmentation model weights are loaded and then frozen to maintain their best performance and stabilize the training process. It is then used to segment the raw motion data $x$ and retrieve sub-action labels $s$. 

To further capture the semantic relationship between sub-actions and holistic actions, we employ the pre-saved label map, denoted by $\mathscr{T}$ to retrieve text labels of holistic actions. $\mathscr{T}_s$ is the equivalent function for sub-action to retrieve the textual description $s_{text}$ corresponding to the encoded sub-action label $s$ (for instance, from one-hot encoded class ``654'' to ``walk straight forward'') as follows:
\begin{equation}
s_{text} = \mathscr{T}_s(s),
\end{equation}
Text labels are encoded into a latent space through tokenization. A followed text encoder is used to standardize and normalize them. We first convert the sub-action text into token indices $t$ as follows:
\begin{equation}
t = Tokenizer(s_{text}).
\end{equation}
We construct the text encoder following the encoder design of CLIP~\cite{CLIP}, including $E$ as the embedding layer to embed the token indices $t$, a classic transformer architecture, and the normalization in the end. Tokens are passed to the text encoder to extract the sub-action label text feature $T_{sub}$ as:
\begin{equation}
    T_{sub} = Norm(Transformer(E(t))).
\end{equation}

\subsection{Cross-Modal Fusion}

To fuse motion and text features for the final classification, we apply a cross-attention mechanism to skeleton-based features $X^{(N)}$ extracted by GCN module and the sub-action text embeddings $T_{sub}$ extracted by the sub-action semantics branch. First, we project them into $Q$ (query), $K$ (key) and $V$ (value) spaces:
\begin{equation}
    Q=X^{(N)} W_Q,  
\end{equation}
\begin{equation}
    K=MLP_1(T_{sub}) W_K,  
\end{equation}
\begin{equation}
    V=MLP_1(T_{sub}) W_V, 
\end{equation}
where $W_Q$, $W_K$, and $W_V$ are embedding weights, and $MLP_1$ is an multilayer perceptron (MLP) with batch normalization, ReLU, and dropout layers. We use $MLP_1$ to apply transformations to the sub-action semantic feature and make its shape match the skeleton-based feature. Especially, stretching the context length to match the sequence length dimension of $X^{(N)}$. The cross-modal fusion is defined as:
\begin{equation}
X_{fused} = Norm(X^{N} + softmax(\frac{QK^{T}}{\sqrt{d_k}})V),
\end{equation}
where $d_k$ is the dimension of $K$ and $Q$. The fused feature $X_{fused}$ is computed along with a residual connection and normalization to stabilize training. Finally, average pooling denoted by $pool_a$ and a linear layer is applied for final action class prediction $\hat{y}$:
\begin{equation}
\hat{y} = linear(pool_a(X_{fused})).
\end{equation}

\subsection{Semantic-Aware Multi-Task Learning}
We design the semantic loss function to enforce the semantic consistency between holistic actions and sub-actions, and also help the text encoder to learn effectively on semantic feature extraction. First, similar to the pipeline of the sub-action text label feature $T_{sub}$, we obtain text feature $T_{hol}$ of the holistic action class $y$ by:
\begin{equation}
T_{hol} = Norm(Transformer(E(Tokenizer(\mathscr{T}(y))))).
\end{equation}
We then compute the semantic loss $\mathcal{L}_{semantic}$ between $T_{sub}$ and $T_{hol}$ using cosine similarity: 
\begin{equation}
    \mathcal{L}_{semantic} = 1 - \frac{MLP_2(T_{sub})\cdot T_{hol}}{\left\| MLP_2(T_{sub}) \right\| \left\| T_{hol} \right\|},
\end{equation}
where $MLP_2$ is an MLP with ReLU and dropout to merge sub-action features and apply transformation to feature dimensions, and $y$ is the ground truth action label. The overall loss combines the semantic loss $\mathcal{L}_{semantic}$ and the cross-entropy loss $\mathcal{L}_{recog}$ between $y$ and $\hat{y}$ from action recognition:
\begin{equation}
    \mathcal{L} = \lambda_1 \mathcal{L}_{recog} + \lambda_2 \mathcal{L}_{semantic},
\end{equation}
where $\lambda_1$ and $\lambda_2$ here are hyperparameters to balance the weights of each loss. These two loss functions work corporately in a multi-task learning manner to recognize the motion patterns and learn the relationship between action and sub-action in the semantic space.

For $MLP_1$, since the purpose is to extract semantic features and fuse with the skeleton-based features, we apply multiple ReLU activations with batch normalization at first. However, for $MLP_2$, we only use the linear layers and dropout to avoid affecting the text embedding ranges. 

\section{Experiment}
\subsection{Dataset Cleansing}
\begin{figure}[th!]
  \centering
  \setlength{\abovecaptionskip}{2pt} 
  \includegraphics[width=0.90\linewidth]{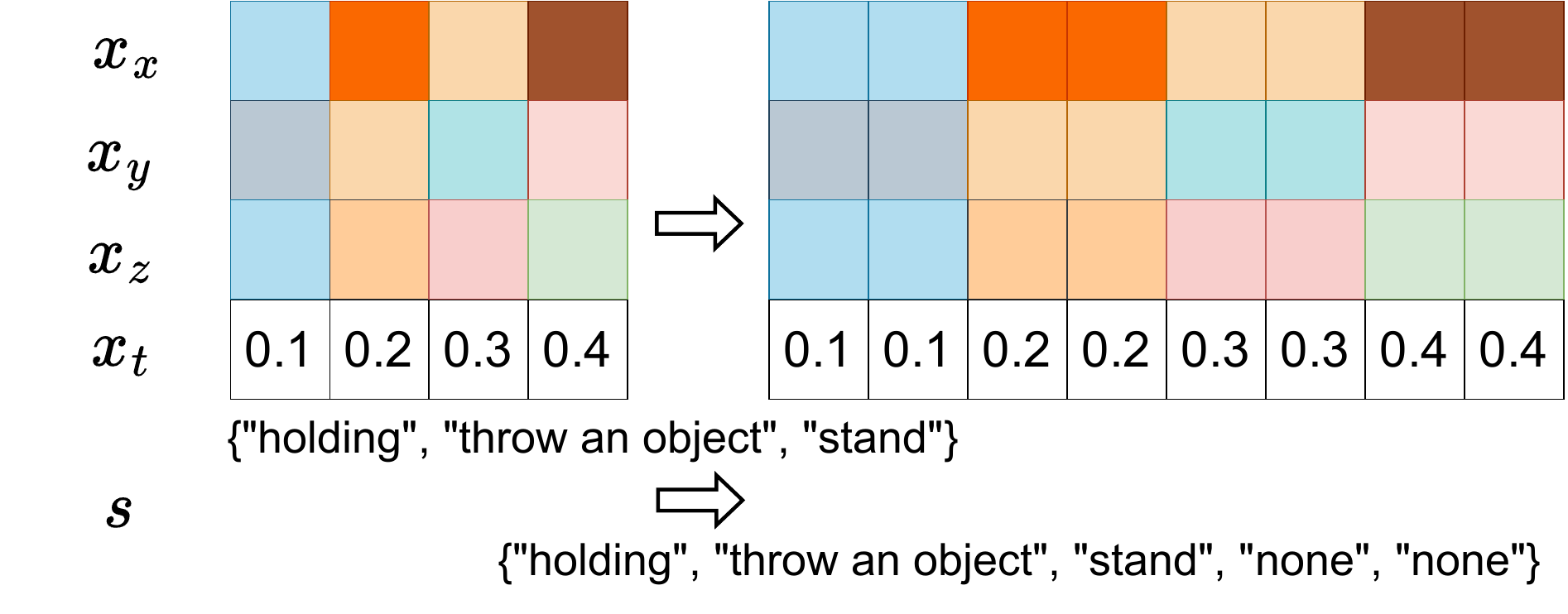}
  \caption{Illustration of data interpolation.}
  \label{fig:padding}
\end{figure}
We evaluate our proposed model SASI and state-of-the-art methods on the BABEL dataset, from which we introduce its sub-action annotations to motion sequences from AMASS dataset~\cite{AMASS:ICCV:2019}. BABEL’s raw labels suffer from redundancy and noise due to multi-dataset aggregation. To address this, we apply the strategy proposed in TEMOS~\cite{temos} to merge similar labels by comparing the cosine similarity using the pre-trained language model. In our implementation, we choose to use the pre-trained Sentence-BERT~\cite{sbert} model as it is demonstrated to be more suitable to the task~\cite{humantomato}. We first compute the similarity score between each label, then group similar labels and treat labels in the same group as the same labels. For the input data, we append timestamps as a feature to preserve temporal continuity. As shown in~\cref{fig:padding}, to standardize the dataset format, joint positions $x$ (including $x_x$, $x_y$, $x_z$, and timestamps $x_t$) are interpolated to a fixed sample length using the nearest-neighbor strategy. For sub-action label sequences, we pad ``none'' tokens at the end and align them to the length of the longest sample to maintain consistency.

\begin{table*}[thp!]
  \vspace{2mm}
  \caption{Performance comparison at different observation ratios (OR)}
  \label{tab:TMP}
  \begin{tabularx}{\linewidth}{lc*{8}{>{\centering\arraybackslash}X}}
    \toprule
    Method & Publication & Param.  & \multicolumn{4}{c}{Separately Trained on each OR (\%)} & \multicolumn{3}{c}{Trained on 100\% OR (\%)} \\
    \cmidrule(lr){4-7} \cmidrule(lr){8-10} 
    & & & 25\% & 50\% & 75\% & 100\% & 25\% & 50\% & 75\%  \\
    \midrule
    Skeleton \\
    \midrule
    ST-GCN~\cite{stgcn} & AAAI 2018 & 3.17M &  46.91 & 53.65 & 54.94 & 57.71 & \textbf{37.21 }& 51.62  & 55.49\\
    2s-AGCN~\cite{2sagcn2019cvpr} & CVPR 2019 & 3.54M & 41.37 & 47.92 & 47.37 & 52.08 & 18.55 & 35.00 & 45.61 \\
    Shift-GCN~\cite{cheng2020skeleton} & CVPR 2020 & \underline{0.81M} & 48.85 & 55.49 & 57.80 & 57.53 & 32.41 & 51.25 & 56.14\\
    CTR-GCN~\cite{CTR-GCN}& ICCV 2021 & 1.49M  &43.77 & 54.57 & 54.20 & 58.17 & \underline{32.87} & 51.71 & 56.05 \\
    Mixformer~\cite{skeletonmixformer}& MM 2023 & 3.85M   & 47.74 & 53.74 & 55.68 & 57.98 & 26.96 & 44.23 & 50.32 \\        
    InfoGCN++~\cite{infogcn++}& TPAMI 2024 & \textbf{0.72M} & 44.51 & 49.22 & 49.05 & 55.96 & 30.56 & 47.83 & 55.22 \\
    BlockGCN~\cite{blockgcn}& CVPR 2024 & 1.89M  & 49.58 & 55.77 & 57.43 & 58.82 & 29.09 & 50.69 & 57.34 \\
    Skateformer~\cite{skateformer} & ECCV 2024 & 3.33M & 41.18 & 49.22 & 52.35 & 56.33 & 31.67 & 46.63 & 49.95   \\
    DeGCN~\cite{degcn}& TIP 2024 & 1.53M & \underline{51.06} & 54.48 & 55.31& 58.08 & 32.59 & \underline{52.35} & 55.31 \\
    ProtoGCN~\cite{protoGCN} & CVPR 2025 & 4.30M & 48.20 & 53.74 & 57.43 & 58.08 & 28.62 & 48.48 & 56.05 \\
    \midrule
    Skeleton + text \\
    \midrule
    GAP~\cite{xiang2023generative} + CTR-GCN  & ICCV 2023 & 2.18M & 43.84 & 52.56 & 53.79 & 56.53 & 31.82 & 51.80 & 56.34\\
    SASI + CTR-GCN &  & 56.60M & 49.40 & 55.68 & 56.14 & \underline{59.10} & 29.36 & 52.08 & \underline{57.43} \\
    SASI + BlockGCN & & 56.99M & 50.14 & \underline{56.42} & \textbf{58.54} & \underline{59.10} & 30.38 & 51.43 & 57.25  \\
    SASI + ProtoGCN &  & 55.56M & \textbf{51.71} & \textbf{56.97} & \underline{58.08} & \textbf{59.65} & 31.76 & \textbf{53.92} & \textbf{57.62}  \\
  \bottomrule
  \end{tabularx}
\end{table*}
\subsection{Implementation}

For the segmentation model, we adopt the state-of-the-art action segmentation model called FACT~\cite{lu2024fact}. Although the original FACT works on video features, we modify it by replacing video features with skeleton-based features to fit the MoCap data and find it also works with skeleton-based human action segmentation. We perform segmentation model training using the same dataset and the same splits mentioned previously. The segmentation model weights are frozen during the training process of SASI. When training the FACT model, we set the action tokens of the action branch to the shape of the padded sub-actions in our dataset to ensure that the output sub-action amount matches the processed dataset. The rest of the hyperparameters are kept the same as the original model. In our implementation, we use the sub-action label outputs without action boundaries.

SASI achieves an inference speed of approximately 29 Hz on an NVIDIA RTX 4090 GPU with an input sequence length of 500, without applying model compilation or optimization techniques.

\subsection{Evaluation}

We compare the proposed method with state-of-the-art methods on the BABEL benchmark, and the results are listed in~\cref{tab:TMP}. We use bold font for the best results and underline the sencond-best results. The pre-trained segmentation model parameters are not included in Param. All trainings are conducted using the same set of hyperparameters. Only joint data are used as input for the testing phase. To assess the improvement introduced by our method on state-of-the-art models, we use CTR-GCN, BlockGCN, and ProtoGCN individually as GCN backbones for skeleton-based feature extraction. Similar to \cite{infogcn++}, we introduce the observation ratio (OR) to evaluate the performance on the partial dataset, where we set ${25\%, 50\%, 75\%, 100\%}$ ORs for independent training and evaluation, and also evaluate the model trained on $100\%$ OR using test sets in ${25\%, 50\%, 75\%}$ ORs. Results show that the proposed method outperforms other methods on independent trained experiments of ${25\%, 50\%, 75\%, 100\%}$, indicating improvements both in complete action recognition and partial action recognition. This result further confirms that sub-action semantics are essential to the action understanding in dynamic environments. Besides, the proposed method trained on $100\%$ OR achieves superior accuracy on test sets in ${50\%, 75\%}$ ORs, suggesting the generalizability of our sub-action-aware learning strategy on partial actions. This illustrates that the learned hierarchical relationships are transferable to partial observations, a key requirement for real-time robotic perception system. Our model's high accuracy at low ORs (25-75\%) demonstrates a crucial low-latency perception capability, enabling a robot system to shift from a reactive to a proactive stance for safe and efficient responses in collaborative tasks. Furthermore, despite the increase in parameter size, augmenting state-of-the-art models with SASI consistently improves performance across most ORs, demonstrating that SASI functions as a modular component.

Moreover, we evaluate the segmentation accuracy of the pre-trained FACT segmentation model used in our framework using the same observation metric. The segmentation accuracies across different ORs are presented in~\cref{tab:segaccuracy}. We evaluate the per-action segmentation accuracy—defined as the proportions of sub-actions correctly recognized out of the total number of them.
\begin{table}[h!]
  \caption{The pre-trained FACT model accuracy at the action level}
  \begin{tabularx}{\linewidth}{l*{4}{>{\centering\arraybackslash}X}}
    \toprule
    Metric & 25\% & 50\% & 75\% & 100\%  \\
    \midrule
    & \multicolumn{4}{c}{Model Separately Trained on each OR (\%)}\\
    \cmidrule(lr){2-5}
    Action & 39.88 & 40.76 & 42.04 & 49.02 \\
    \midrule
    & \multicolumn{4}{c}{Model Trained on 100\% OR (\%)} \\
    \cmidrule(lr){2-5}
    Action & 35.70 & 44.03 & 48.08 & \\
    \bottomrule
  \end{tabularx}
  \label{tab:segaccuracy}
\end{table}
\subsection{Ablation Study}
To evaluate the contributions of different modules to the performance improvement of the proposed model, we conduct a comprehensive ablation study by removing or replacing modules to analyze their effects on the model performance. All experiments are performed on the same benchmark with identical hyperparameters. Similar to previous multi-modal methods \cite{xiang2023generative, liu2024multi}, we adopt CTR-GCN as the GCN backbone for the ablation studies.

\noindent
{\bf Effectiveness of Sub-Action Segmentation Enhancement}
We evaluate the impact of sub-action segmentation on human action recognition. To analyze the effect of different segmentation accuracies, we replace the segmentation model output with ground truth sub-action labels and inject $80\%, 60\%, 40\%, 20\%$ random errors to simulate segmentation accuracies of $20\%, 40\%, 60\%, 80\%$, respectively. This error is applied at the action level to preserve the continuity of segmentation outputs. As a result, both the number and class of sub-actions in each sample may deviate from the ground truth. 
We report the mean and standard deviation of four accuracy metrics over three runs with different random seeds. The results in~\cref{tab:segablation} show that segmentation accuracy above 40\% consistently improves recognition performance. Particularly, the 100\% segmentation model yields a significant gain (10.99\%), supporting the hypothesis that sub-action segmentation benefits skeleton-based action understanding. Furthermore, recognition accuracy increases monotonically with segmentation quality, indicating a strong positive correlation.

These results provide an important insight: the overall recognition performance is bounded by the quality of sub-action segmentation. While the current segmentation model has limited accuracy, improved segmentation consistently leads to better recognition results, indicating that better segmentation can further improve the performance of SASI.

However, we note that simulated segmentation accuracy cannot fully reflect real conditions, as random error cannot capture class-wise error distributions in practical cases.
\begin{table}[h]
  \centering
  \caption{Ablation study on the segmentation accuracy}
  \begin{tabular}{ccc}
    \toprule
    Segmentation Acc. (\%) & Mean Acc. (\%) & SD (\%)\\
    \midrule
    Baseline & 58.17 & --- \\
    20\% & 57.83 & 0.42\\
    40\% & 57.74 & 0.43 \\
    60\% & 59.89 & 2.15\\
    80\% & 65.13 & 4.54 \\
    100\% & 69.16 & --- \\
    \bottomrule
  \end{tabular}
  \label{tab:segablation}
\end{table}

\noindent
{\bf Effectiveness of Cross-Attention Fusion}
\begin{figure*}[th!]
  \centering
  \setlength{\abovecaptionskip}{2pt} 
  \includegraphics[width=0.92\linewidth]{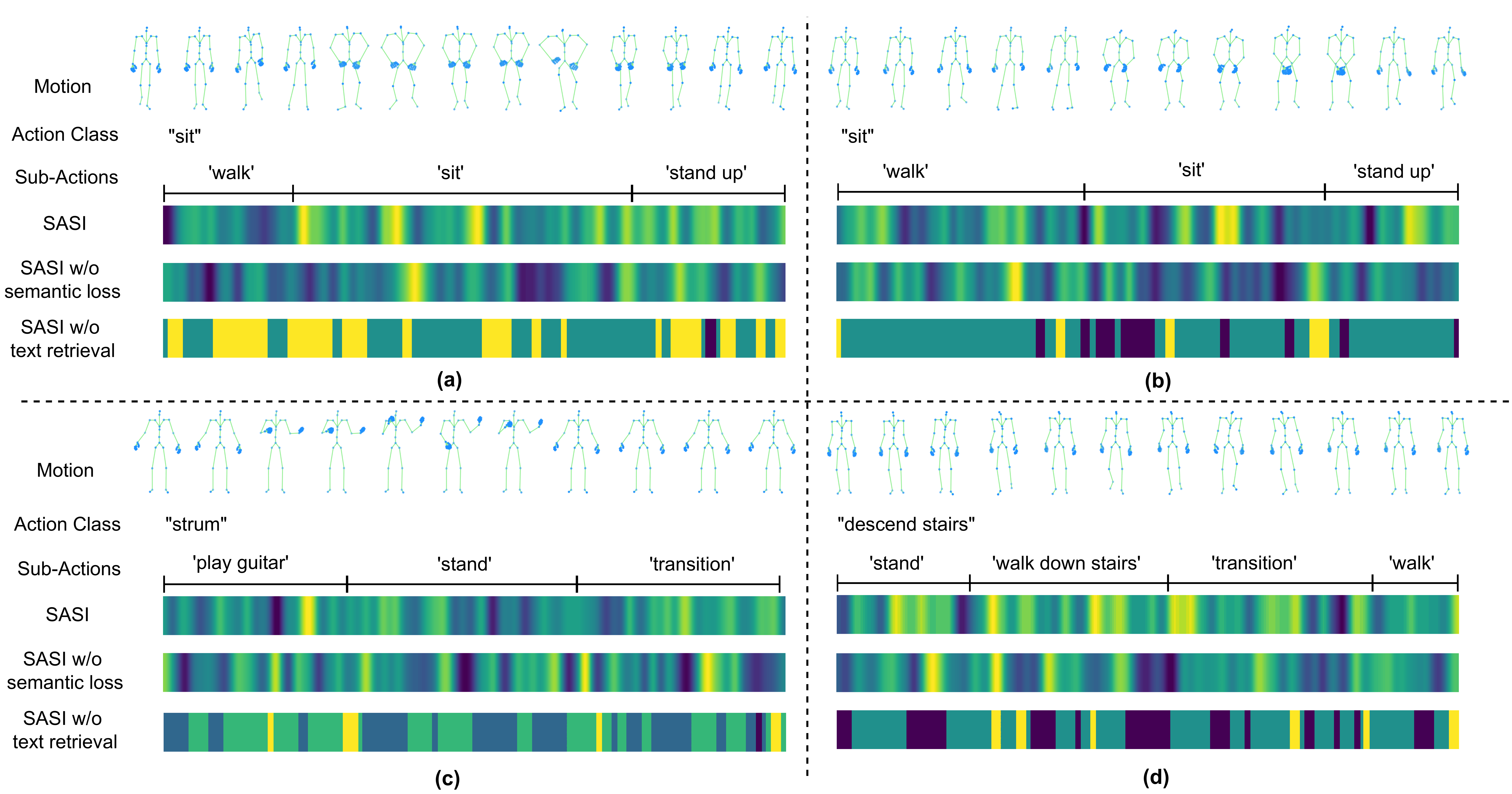}
  \caption{Visualization of attention weights in the cross-modal fusion module for four sample sequences comparing the cross-attention outputs of the complete SASI model, SASI without semantic loss, and SASI without text retrieval.}
  \label{fig:visualization}
\end{figure*}
We compare our cross-modal attention with the other fusion methods on the cross-modal fusion module: element-wise addition, element-wise multiplication, and concatenation. Moreover, we add the experiment of element-wise multiplication with residual connection and the experiment of cross-attention without residual connection. As shown in~\cref{tab:fusion_ablation}, cross-attention outperforms the rest of the methods. The accuracy of cross-attention without residual connection is 0.56\% lower than the best accuracy, but still larger than others, indicating the advantage of the cross-attention mechanism in cross-modal feature fusion tasks. 

As shown in~\cref{fig:visualization}, we conduct qualitative analysis on the cross-attention module. Attention weights computed by $softmax(\frac{QK^{T}}{\sqrt{d_k}})$ in the attention mechanism, are projected into a 1D graph to visualize the attention in the motion sequence. Light areas (yellow) represent high attention scores and dark areas (purple) represent low attention scores. Additionally, the holistic action, sub-actions, and human motion are attached. Generally, we can observe that SASI’s attention scores are higher in key sub-action regions, indicating that the proposed method emphasizes the semantic relationship between the holistic action and its critical sub-actions. For instance, in~\cref{fig:visualization} (a), sub-action ``sit'' has a higher attention score than other sub-actions, showing its essential association with its parent action ``sit''. Sample in~\cref{fig:visualization} (b) with similar sub-actions further confirms the effectiveness. Moreover,~\cref{fig:visualization} (c) shows that the sub-action ``play guitar'' is the most highlighted area in action ``strum''.~\cref{fig:visualization} (d) does not have a prominent attention score in one specific sub-action, but shows a broad focus around ``walk down stairs'' area. These results reveal that attention weights focus on task-relevant regions, confirming the importance of dynamically modeling inter-modal dependencies. 

\begin{table}[h]
\centering
\caption{Ablation study on feature fusion strategies}
\label{tab:fusion_ablation}
\centering
\begin{tabular}{cc}
    \toprule
    Fusion Method       & Acc. (\%)  \\
    \midrule
    Element-wise Addition    & 53.09\\
    Element-wise Multiplication w/o residual& 52.82\\
    Element-wise Multiplication w/ residual &  54.85\\
    Concatenation & 54.94\\
    Cross-Attention w/o residual & 57.89 \\
    Cross-Attention w/ residual & 59.10 \\
    \bottomrule
\end{tabular}
\end{table}
\noindent
{\bf Effectiveness of Semantic Loss Function}
Next, we analyze the contribution of the proposed semantic loss function which compares the similarity between predicted sub-actions and the holistic action. We compare the recognition accuracy of our proposed method with and without the action recognition loss $\mathscr{L}_{recog}$ to evaluate its effectiveness. Results in~\cref{tab:lossablation} show that the recognition accuracy is better when the semantic loss is applied, confirming that the semantic loss is essential to reveal the correlation between action and sub-actions thus improving model performance. Furthermore, we visualize the attention weights in the cross-modal fusion module for several samples to illustrate the effect of semantic loss. Results in~\cref{fig:visualization} show that SASI trained without semantic loss cannot grasp the essential semantic relationship between holistic actions and sub-actions. For instance, in~\cref{fig:visualization} (c), the network pays more attention to the end of the sequence which is a static transition action that is comparatively irrelevant to the holistic action ``strum''. Instead, SASI with the semantic loss pays more attention to the ``play guitar'' sub-action. 

\begin{table}[h!]
  \centering
  \caption{Ablation study on the semantic loss function and the text retrieval function}
  \begin{tabular}{cc}
    \toprule
    Method & Recognition Acc. (\%) \\
    \midrule
    w/ semantic loss function \& text retrieval & 59.10 \\
    w/o semantic loss function & 57.43 ($\downarrow$ 1.67) \\
    w/o text retrieval  & 52.82 ($\downarrow$ 6.28)  \\
    \bottomrule
  \end{tabular}
  \label{tab:lossablation}
\end{table}

\noindent
{\bf Effectiveness of Text Retrieval}
We evaluate the essentialness of text retrieval and the text encoder in the sub-action semantics branch. In~\cref{tab:lossablation}, we compare the performance of the proposed model with and without text retrieval. In the variant without it, the segmented sub-action classes are directly represented as one-hot encoded vectors and fed into the encoder without undergoing text retrieval and tokenization. Results show that text retrieval improves the accuracy by 6.28\%, demonstrating its crucial role in leveraging text labels to extract semantic information from human actions for the exploration of the semantic relationships between sub-actions and holistic actions. Moreover,~\cref{fig:visualization} shows the attention weight plots for SASI without text retrieval. Compared to the complete SASI model, these attention weights are noticeably simpler, indicating that the cross-modal fusion process without text retrieval fails to capture the complexity of semantic information.
\section{Conclusion}
In this work, we proposed SASI to address two key challenges: the intrinsic complexity of real-world actions and the difficulty of understanding incomplete motion sequences. By integrating fine-grained sub-action semantics with spatiotemporal features extracted by a graph convolution network, SASI effectively captures the hierarchical structure inherent in human actions. Experimental results, particularly the strong performance on partial sequences, demonstrate its effectiveness in enabling more proactive, safe, and efficient HRI. As a modular component to enhance state-of-the-art backbones, SASI offered a practical pathway to embedding a semantic, hierarchical understanding of human actions into future robotic systems. An important observation from our experiments is that the overall performance is bounded by the quality of the sub-action segmentation module. Although jointly designed within our framework, its current accuracy limits recognition performance, while improvements in segmentation consistently yield better results.

\noindent
{\bf Limitations and Future Work} Our approach has several limitations that suggest possibilities for future work. First, the overall performance is constrained by the relatively low accuracy of the pre-trained segmentation model, which limits the upper bound of the overall system performance. Second, the dependency on sub-action annotations restricts the method's application. Third, sub-action annotations can be refined to represent detailed body states, rather than solely in sequential order. Finally, the large number of sub-action classes makes the dataset insufficient for robust sub-action learning. Future work could address these challenges by exploring end-to-end joint training of the segmentation and recognition modules for better hierarchy modeling. Moreover, developing unsupervised techniques to discover action primitives from samples would broaden the method applicability.

\section{Acknowledgement}
This research is partially supported by Initiative on Recommendation Program for Young Researchers and Woman Researchers, Information Technology Center, The University of Tokyo.

\bibliographystyle{IEEEtran}
\bibliography{main}

\end{document}